\documentclass[runningheads]{llncs}
\usepackage{graphicx}

\usepackage{tikz}
\usepackage{comment}
\usepackage{amsmath,amssymb} 
\usepackage{color}
\usepackage{booktabs}
\usepackage{multirow}
\usepackage{yhmath}
\usepackage{amsmath}
\usepackage{array}

\newcommand{\down}[1]{\tiny\textcolor{red}{#1}}
\newcommand{\up}[1]{\tiny\textcolor{blue}{#1}}
\usepackage[accsupp]{axessibility}  

\makeatletter
\def\thanks#1{\protected@xdef\@thanks{\@thanks
        \protect\footnotetext{#1}}}
\makeatother

\begin{document}
\pagestyle{headings}
\mainmatter
\def\ECCVSubNumber{798}  

\title{Learning with Recoverable Forgetting} 

\titlerunning{ECCV-22 submission ID \ECCVSubNumber} 
\authorrunning{ECCV-22 submission ID \ECCVSubNumber} 
\author{Anonymous ECCV submission}
\institute{Paper ID \ECCVSubNumber}

\titlerunning{Learning with Recoverable Forgetting}
%
\author{Jingwen Ye\inst{1}
\and
Yifang Fu\inst{1}\and
Jie Song\inst{2}
\and
Xingyi Yang\inst{1}
\and
Songhua Liu\inst{1} \and
Xin Jin \inst{3}  \and
Mingli Song \inst{2} 
\and
Xinchao Wang \inst{1}$^{\dagger}$ \thanks{ $^{\dagger}$ Corresponding author.}
}

\authorrunning{J. Ye et al.}
%
\institute{National University of Singapore\and
Zhejiang University \quad \textsuperscript{\rm 3} Eastern Institute of Advanced Study\\
\email{\{jingweny,xinchao\}@nus.edu.sg},
\email{\{e0724403,xyang,songhua.liu\}@u.nus.edu}
\email{\{sjie,brooksong\}@zju.edu.cn},
\email{jinxin@eias.ac.cn}}

\maketitle

\begin{abstract}
Life-long learning aims at learning a 
sequence of tasks without forgetting the previously acquired knowledge.
However, the involved training data may not be life-long
legitimate due to privacy or copyright reasons.
In practical scenarios, for instance, 
the model owner may wish to enable or disable
the  knowledge of specific tasks or specific samples
from time to time.
Such flexible control over knowledge transfer, unfortunately,
has been largely overlooked
in previous incremental or decremental learning methods,
even at a problem-setup level.
In this paper, we explore a novel 
learning scheme, termed as 
\textbf{L}earning w\textbf{I}th \textbf{R}ecoverable \textbf{F}orgetting (LIRF),
that explicitly  handles 
the task- or sample-specific knowledge removal and recovery.
Specifically, 
LIRF brings in two innovative 
schemes, namely knowledge \emph{deposit} and \emph{withdrawal},
which allow for isolating user-designated 
knowledge from a pre-trained network
and injecting it back when necessary.
During the knowledge deposit process,
the specified knowledge is extracted
from the target network and stored in a deposit module,
while the insensitive or general knowledge of the 
target network is preserved and  further augmented.
During knowledge  withdrawal, 
the taken-off knowledge is added back to the target network.
The deposit and withdraw processes 
only demand for a few epochs of finetuning on the 
removal data, ensuring both data and time efficiency. 
We conduct experiments on several datasets,
and deomnstrate that the proposed LIRF
strategy yields encouraging results
with gratifying generalization capability.

\end{abstract}

\section{Introduction}
Life-long learning finds its application
across a wide spectrum of domains and 
has been a long-standing research task.
Its main goal is to update a network
to adapt to new data, such as
new instances or samples from a new class,
without forgetting the learned knowledge on the past data.
In some scenarios, on the contrary,
we wish to deliberately
forget or delete specified knowledge
stored in the model, due to privacy
or copyright issues.
This task, known as, machine unlearning,
has also attracted increasing attentions
from the industry and research community
due to its practical value.

Nevertheless, prior attempts in machine unlearning have been
mostly focused on deleting the specified knowledge for good,
meaning that once removed, it is not 
possible to revert the knowledge
back. Despite the absolute IP protection, such
knowledge deletion scheme indeed introduces 
much inconvenience in terms of the user control
and largely hinders the flexibility of model interaction.

In this paper, we explore a novel learning scenario,
which explicitly allows for the extracted knowledge
from a pre-trained networked to be deposited and, whenever needed, 
injected back to the model. 
Such a flexible learning strategy
grants users a maximum degree of freedom
in terms of control over task- or sample-specific
knowledge, and meanwhile 
ensures the network IP protection.
Admittedly, this ambitious goal 
inevitably leads to a more challenging
problem to tackle, since again
we seek a portable modulation 
of knowledge on and off 
a pre-trained network.

\begin{figure}[t]
\centering
\includegraphics[scale = 0.42]{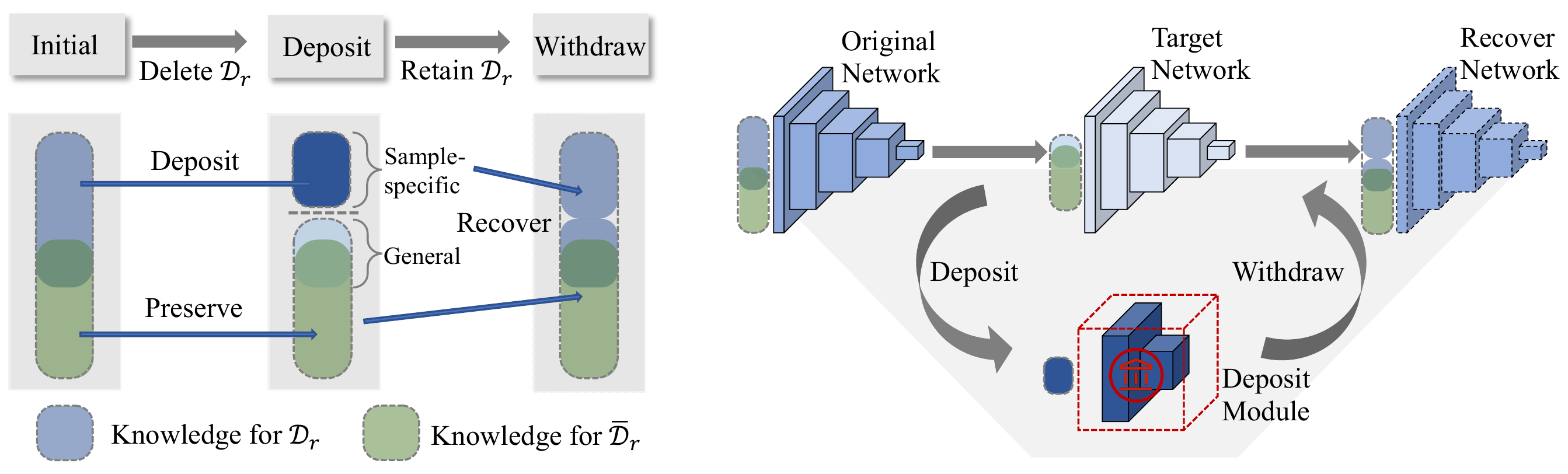}
\caption{Illustration of the proposed LIRF framework, comprising the knowledge deposit process and knowledge withdrawal process.
}
\label{fig:goal}
\end{figure}

To this end, we propose a dedicated scheme, termed as
Learning with Recoverable Forgetting (LIRF).
We illustrate the overall pipeline of LIRF in Fig.~\ref{fig:goal},
When there is a request for deleting specified knowledge, denoted
as $\mathcal{D}_r$ (with $\overline{\mathcal{D}}_r$ preserved), 
due to for example IP issues, 
LIRF isolates such knowledge from the pre-trained 
original network
and stores it in a \emph{deposit module};
the remaining network with $\mathcal{D}_r$ 
extracted is then denoted as the \emph{target network}.
When the IP issue is resolved and the model owner requests
to revert the knowledge back or re-enables $\mathcal{D}_r$, LIRF withdraws
the deposited knowledge and amalgamates it
with the target network
to produce the \emph{recover network}.
Specifically, during the knowledge deposit process,
we partition the knowledge of the original networks, 
trained using full data, into sample-specific and general part.
The former is deposited to a deposit module consisting 
of pruned blocks from the original network,
while the latter is preserved in the target network.

Our contributions are therefore summarized as follows.
\begin{itemize}
    \item We introduce a novel yet practical life-long learning setup, 
    recoverable knowledge forgetting.
    In contrast to  machine unlearning settings
    that delete specified knowledge for good,
    recoverable forgetting 
    enables  knowledge isolation and recovery 
    from a pre-trained network, 
    which brings in network IP protection alongside
    user flexibility and control.
    
    \item We develop the LIRF framework
    that explicitly allows for knowledge deposit
    and withdrawal, to achieve 
    recoverable knowledge forgetting.
    LIRF is time- and data-efficient,
    as the deposit process requires only
    a few epochs to finetuning  on the 
     specified samples.
   
     \item Experimental results have verified the effectiveness of the proposed method, under various settings including 
     class-incremental learning and machine unlearning.
\end{itemize}

\section{Related Work}

\subsection{Life-long Learning}

Life-long/online/incremental learning, which is capable of learning, retaining and transferring knowledge over a lifetime, has been a long-standing research area in many fields~\cite{Wu2018MemoryRG,Shmelkov2017IncrementalLO,Huihui21AAAI}. 
As the pioneer work, Li~\textit{et al.}~\cite{Li2016LearningWF} propose Learning without Forgetting (LwF) by using only the new-coming examples for the new task's training, while preserving the responses on the existing tasks to prevent catastrophic forgetting. 
Peng \textit{et al.}~\cite{Peng2017IncrementallyLT} present to train the hierarchical softmax function for deep language models for the new-coming tasks. FSLL~\cite{Mazumder2021FewShotLL} is proposed to perform on the few-shot setting by selecting very few parameters from the model.
Apart from those works that still need part of the old data, many researchers are devoted to developing the methods without storing the old data by synthesizing old data~\cite{Choi2019AutoencoderBasedIC,Shin2017ContinualLW,Venkatesan2017ASF} or even without referring to any old data~\cite{Sun2018ActiveLL,Shmelkov2017IncrementalLO,Nekoei2021ContinuousCA}.
In addition to the above works that tend to forbid the catastrophic forgetting of the old tasks, some researchers~\cite{Hou2018OnePassLW,Zhang2020LearningWF,Hou2021LearningWF,Hou2021StorageFL} pay more attention on the decremental cases where some features may vanish while feature evolving. Hou \textit{et al.}~\cite{Hou2018OnePassLW} attempt to compress important information of vanished features into functions of survived features, and then expand to include the augmented features in the one-pass learning way. Zhang \textit{et al.}~\cite{Zhang2020LearningWF} propose discrepancy measure for data with evolving feature space and data distribution.

Different from the current researches on life-long learning, we propose the more flexible learning scheme, which is capable of dealing with both the data adding and deleting cases.

\subsection{Knowledge Transfer}
Knowledge transfer aims at transferring knowledge from networks to networks. Here, we mainly discuss the related works in knowledge distillation~\cite{hinton2015distilling,Han2020NeuralCM,yang2020CVPR}, which trains a student model of a compact size
by learning from a larger teacher model or a set of teachers handling the same task. It has been successfully conducted in deep model compression~\cite{WangCVPR17}, incremental learning~\cite{Rosenfeld2020IncrementalLT}, continual learning~\cite{Lange2021ACL,ye2022safe} and other tasks other than classification~\cite{Chen2017LearningEO,yang2020NeurIPS,Huang2018KnowledgeDF,Xu2018PADNetMG,Sucheng2022CVPR,Weihao22MetaFormer,YujingCVPR22,JingwenCVPR22,jing2020dynamic}.
In addition to the above methods that transfer knowledge from one network to another, it could happen in plenty forms.
Such as for combining or amalgamating multi-source knowledge,
Gao \textit{et al.}~\cite{gao2017knowledge} introduce a multi-teacher and single-student knowledge concentration approach. And in order to handle multi-task problems in one single network, knowledge amalgamation~\cite{ye2019student} is proposed to train the student network on multiple scene understanding tasks, leading to better performance than the teachers. To make it further, Ye \textit{et al.}~\cite{Ye_Amalgamating_2019} apply a two-step filter strategy to customize the arbitrary task set on TargetNet. 
Besides, the multi-stage knowledge transfer is enabled by Yuan \textit{et al.}~\cite{Yuan2020CKDCK} to design a multi-stage knowledge distillation paradigm to decompose the distillation process.

Knowledge distillation could also be a reliable method to transfer knowledge from old data to new data, and there are also some distillation-based works~\cite{Cheraghian2021SemanticawareKD,Hu2021DistillingCE,Tao2020FewShotCL,Dong2021FewShotCL} for solving the coming new data in life-long learning setting.
Cheraghian \textit{et al.}~\cite{Cheraghian2021SemanticawareKD} address the problem of few-shot class incremental learning by utilizing the semantic information. 
Hu \textit{et al.}~\cite{Hu2021DistillingCE} derive a distillation method to retain the old effect overwhelmed by the new data effect, and thus alleviate the forgetting of the old class in testing.

These knowledge transfer methods transfer knowledge from networks to networks, we make the first work to filter and deposit the knowledge.

\subsection{Machine Unlearning}
The concept of unlearning is firstly introduced by Bourtoule \textit{et al.}~\cite{Bourtoule2021MachineU} to eliminate the effect of data point(s) on the already trained model. 
Along this line, Neel \textit{et al.}~\cite{Neel2021DescenttoDeleteGM} give the first data deletion algorithms.
To minimize the retraining time, data removal-enabled forests~\cite{Brophy2021MachineUF} are introduced as a variant of random forests that enables the removal of training data.
Sekhari \textit{et al.}~\cite{sekhari2021remember} initiate a rigorous study of generalization in machine unlearning, where the goal is to perform well on previously unseen datapoints and the focus is on both computational and storage complexity.
Gupta \textit{et al.}~\cite{gupta2021adaptive} give a general reduction from deletion guarantees against adaptive sequences to deletion guarantees against non-adaptive sequences, using differential privacy and its connection to max information. Nguyen \textit{et al.}~\cite{nguyen2020variational} study the problem of approximately unlearning a Bayesian model from a small subset of the training data to be erased. As machine unlearning is studied for data privacy purpose, Chen \textit{et al.}~\cite{chen2021machine} firstly study on investigating the unintended information leakage caused by machine unlearning. 

The previous works only consider the data deletion with the optimization objective of getting the same model as re-training without the deletion data. The proposed LIRF framework only deleted sample-specific knowledge, which can be stored for future use.

\begin{figure}[t]
\centering
\includegraphics[scale = 0.4]{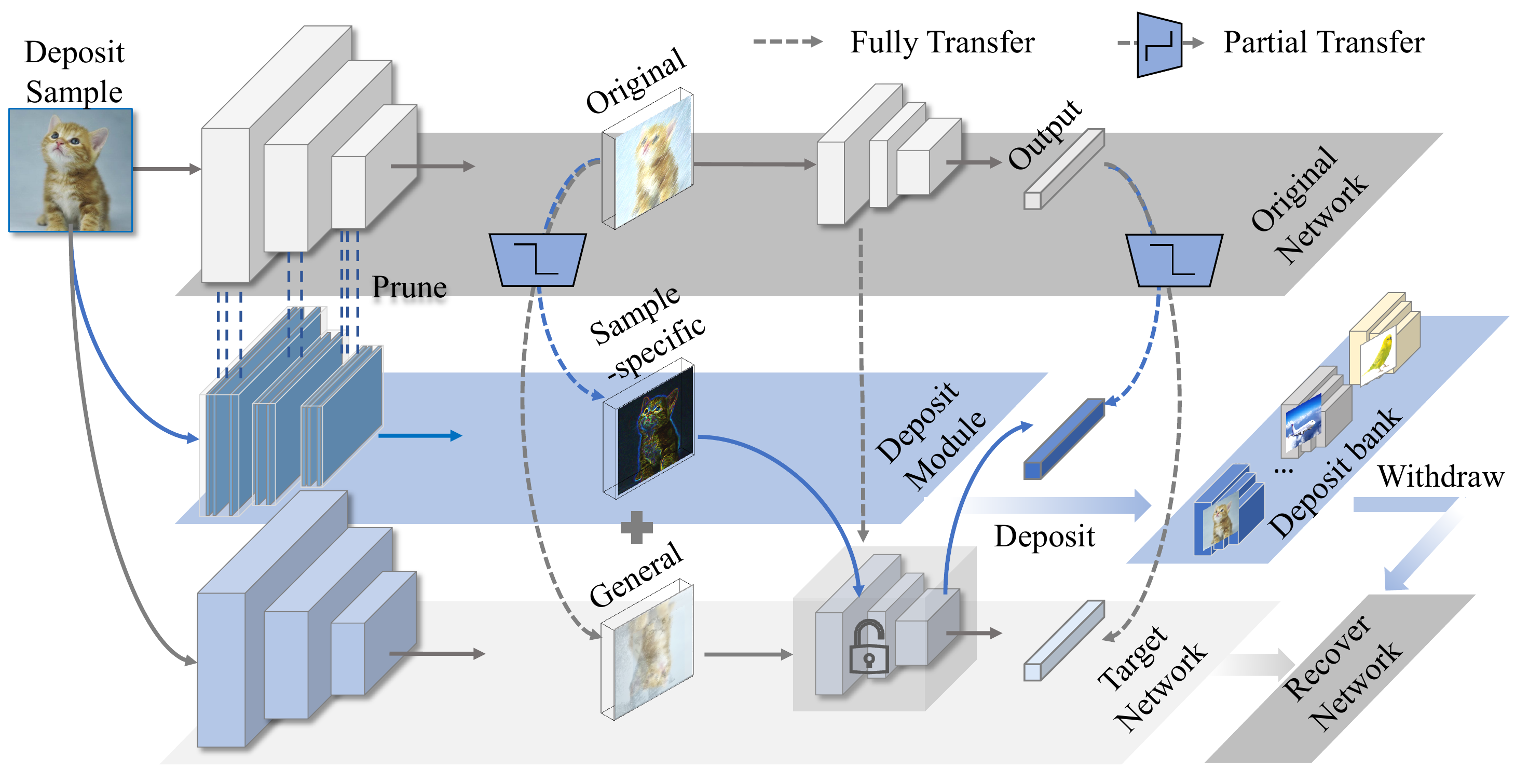}
\caption{The proposed LIRF framework. The knowledge is transferred fully and partially from the original network to the deposit module and the target network. The recover network is withdrawn from the target net and the deposit module.
}
\label{fig:framework}
\end{figure}

\section{Knowledge Deposit and Withdrawal}
The proposed LIRF framework 
focuses on the class-level life-long problem,
in which the samples from multiple classes 
may be deposited or withdrawn.

We define our new problem illustrated in Fig.~\ref{fig:goal}. Let $\mathcal{D}$ be the full original dataset, and the original network directly trained on $\mathcal{D}$ is donated as $\mathcal{T}_0$.
In this problem, each of the learned samples is assigned to either deposit set or preservation set. Formally,
\begin{itemize}
    \item \textbf{Deposit set $\mathcal{D}_r$}: A set of samples that
should be forgotten at the target network $\mathcal{T}$, and remembered at the deposit module $\mathcal{T}_r$;
    \item \textbf{Preservation set $\overline{\mathcal{D}}_r$}: A set of samples that should be memorized at the target network (the complement of $\mathcal{D}_r$).
\end{itemize}
For clarity, we discuss on the case that one deposit set is required for deposit and withdrawal, which could be definitely generalized to multiple deposit sets.

\noindent\textbf{Definition 1} (Deposit Problem). The Learning with knowledge deposit problem is defined as follows:
Learn two models, one is $\mathcal{T}:\mathcal{X}\rightarrow\mathcal{Y}$  that should map an input $x$ to its correct class label $y$ if $x\subset\overline{\mathcal{D}}_r$, while map $x$ to a wrong class label if $x\subset\mathcal{D}_r$; the other one is $\mathcal{T}_r:\mathcal{X}\rightarrow\mathcal{F}$ that stores the knowledge of set $\mathcal{D}_r$. \textit{Constraints}: Only the original network $\mathcal{T}_0$ and deposit set $\mathcal{D}_r$ are available.

\noindent\textbf{Definition 2} (Withdraw Problem). The Learning with knowledge withdraw problem is defined as follows:
Recover a model $\widetilde{{\mathcal{T}}}:\mathcal{X}\rightarrow\mathcal{Y}$  that should map an input $x$ to its correct class label $y$ for both $x\subset{\mathcal{D}}_r$, and $x\subset\overline{\mathcal{D}}_r$. \textit{Constraints}: Only the target network $\mathcal{T}$ and deposit module $\mathcal{T}_r$ are available.

\section{Learning with Recoverable Forgetting}
The essence of this work is to deposit and withdraw the sample-specific knowledge for the deleted data in the learning with recoverable forgetting way, which, we call LIRF framework. LIRF consists of two processes, one is knowledge deposit that transfers knowledge from original network to target network and deposit module, the other is knowledge withdrawal that recovers the knowledge back to the recover net.
These two processes can be described as:
\begin{equation}
    \mathcal{T}_0\xrightarrow[\mathcal{D}_r]{\text{Deposit}} \{\mathcal{T}, \mathcal{T}_r\}\xrightarrow{\text{Withdraw}}  \widetilde{{\mathcal{T}}},
\end{equation}
where $\mathcal{T}_0$ is the original network trained on the full set $\mathcal{D}$, $\mathcal{T}$ is the target network specified for the task of the preservation set $\overline{\mathcal{D}}_r$, $\mathcal{T}_r$ is the deposit module that only works as a knowledge container and $\widetilde{{\mathcal{T}}}$ is the recover network that is expected to recover all the prediction capacity of the full data set $\mathcal{D}$. 

Now, given the original network $\mathcal{T}_0$ and the deposit set $\mathcal{D}_r$,
the goal of LIRF is to learn $\mathcal{T}$, $\mathcal{T}_r$ and $\widetilde{{\mathcal{T}}}$, which includes three steps. Firstly, LIRF filters knowledge out of the original network to get the target net, at the meanwhile, it deposits the filtered sample-specific knowledge to the deposit module, and finally for recover request, LIRF withdraws the knowledge from the deposit module to recover net.
Fig.~\ref{fig:framework} provides an overall sketch of LIRF framework.

\subsection{Filter Knowledge out of Target Net}

In the process of knowledge deposit, the objective of target net is to remove the sample-specific knowledge of $\mathcal{D}_r$ while maintaining the performance on $\mathcal{D}_r$.

To begin with, we divide the original network $\mathcal{T}_0$ into two modules at the $n$-th block, which are denoted as $\mathcal{T}_0^{(-n)}$ and $\mathcal{T}_0^{(n-)}$, respectively. And the target network is divided in the same way as $\mathcal{T}=\mathcal{T}^{(-n)}\circ \mathcal{T}^{(n-)}$.
As has been discussed in the previous work~\cite{Lee2021SharingLI} that upper layers are preferable for transfer in life-long learning setting, $\mathcal{T}_0^{(n-)}$ is fully transferred to the target network.
That is, we fix the last few blocks ($\mathcal{T}^{(n-)}=\mathcal{T}_0^{(n-)}$) and expect this transfer configuration to benefit tasks that share high-level concepts but have low-level feature differences. 
Thus, we fully transfer $\mathcal{T}_0^{(n-)}$ to $\mathcal{T}^{(n-)}$, and partially transfer $\mathcal{T}_0^{(-n)}$ to $\mathcal{T}^{(-n)}$, as the lower layers of the network contain more sample-specific knowledge. 

\subsubsection{Sample-specific knowledge removal.}
This removal is conducted in two aspects.
One is the logit level that the target network is incapable of making reliable prediction on the deposit set $\mathcal{D}_r$, the other is the feature level that the knowledge of $\mathcal{D}_r$ can't be distilled from $\mathcal{T}$. 
Thus, for each input $x\subset \mathcal{D}_r$, we assign a \textit{\textbf{random}} label $y_r$, and force $\mathcal{T}$ to randomly predict on $\mathcal{D}_r$.
And the loss to maximize attention transfer on the intermediate features is applied to the output of $\mathcal{T}^{(-n)}$, which makes $\mathcal{T}$ undistillable for $\mathcal{D}_r$. Thus, the loss $\mathcal{L}_{kr}$ for knowledge removal is calculated as:
\begin{equation}
\mathcal{L}_{kr}= \mathcal{L}_{ce}\big(\mathcal{T}(x),y_r\big)-\lambda_{at} \mathcal{L}_{at}\big(\mathcal{T}^{(-n)}(x),\mathcal{T}_0^{(-n)}(x)\big),
\label{eq:kr}
\end{equation}
where $\lambda_{at}$ is the weight,  $\mathcal{L}_{ce}(\cdot,\cdot)$ is the cross-entropy loss, and $\mathcal{L}_{at}$ is the filtered attention distillation loss item~\cite{zagoruyko2016paying} that calculates the activated feature-wise similarity of the intermediate features:
\begin{equation}
\begin{split}
      \mathcal{L}_{at}(\mathcal{F}_1,\mathcal{F}_2)&= \Big \|f(\frac{A(\mathcal{F}_1)}{\|A(\mathcal{F}_1)\|_2})-f(\frac{A(\mathcal{F}_2)}{\|A(\mathcal{F}_2)\|_2})\Big\|^2,\\
      A(\mathcal{F})&= \sum_{i=1}^C \|\mathcal{F}_i\|^2, 
      \quad f\big(a(i)\big)= \begin{cases}
      0 & a(i)<\epsilon\\
      a(i) & \text{otherwise} \\
      \end{cases},
\end{split}
\end{equation} 
where $\mathcal{F}_i\in\mathbb{R}^{H\times W}$ represents the feature $\mathcal{F}\in\mathbb{R}^{H\times W\times C}$ with the size of $H\times W\times C$ at depth $i$, with which, the $l_2$-normalized attention maps are obtained. And before calculating the attention similarity with $\mathcal{L}_{at}$, a filter function $f$ is applied to set 0 to the deactivate regions, which enables the intermediate knowledge undisillable only for $x\subset\mathcal{D}_r$.   

The knowledge removal loss $\mathcal{L}_{kr}$ is calculated on the deposit set $\mathcal{D}_r$ to fine-tune $\mathcal{T}^{(-n)}$, which is initialized with $\mathcal{T}_0^{(-n)}$. The former loss item of $\mathcal{L}_{kr}$ enables the knowledge forgetting in the logit-level, the latter item of $\mathcal{L}_{kr}$ enables the forgetting in the feature-level, which unlearns $\mathcal{D}_r$ from $\mathcal{T}$ and removes the privacy information of $\mathcal{D}_r$.

\subsubsection{General knowledge preservation.}
As is stated in Fig.~\ref{fig:goal}, there are two kinds of knowledge that need to be preserved by the target network. One is the knowledge coming from the preservation $\overline{\mathcal{D}}_r$, the other is the general knowledge from the $\mathcal{D}_r$. 
Since the target network $\mathcal{T}$ is initialized with the original network and the last few blocks of $\mathcal{T}$ keep fixed while fine-tuning, part of the knowledge has already been preserved by fully transferred from $\mathcal{T}_0^{(n-)}$ to $\mathcal{T}^{(n-)}$.
In addition to it, the partial knowledge transfer with filter $g$ is proposed on the  $\overline{\mathcal{D}}_r$-related knowledge so as to prevent catastrophic forgetting on $\overline{\mathcal{D}}_r$, which is:
\begin{equation}
    \mathcal{L}_{kp} = \mathcal{L}_{kd}\big(g(\frac{z_{\mathcal{T}}(x)}{T}),g(\frac{z_{\mathcal{T}_{0}}(x)}{T})\big),
    \label{eq:kp}
\end{equation}
where $\mathcal{L}_{kd}$ is the KL-divergence loss, $T$ is the temperature, and $z_{\mathcal{T}}$ and $z_{\mathcal{T}_0}$ are the output logits of ${\mathcal{T}}$ and ${\mathcal{T}_0}$, respectively. The filter $g$ selects the logits that correspond to the class of the preservation set, in which way the knowledge is partially transferred to target net by minimizing $\mathcal{L}_{kp}$. Note that only the deposit samples are accessible in the whole LIRF framework, the output probabilities on the preservation class are thought to be low and may not be enough to maintain the performance on the preservation set. Thus, we set a higher temperature weight to transfer more knowledge for the preserved tasks.

\subsection{Deposit Knowledge to Deposit Module}
\label{sec:deposit}
The key difference between the proposed LIRF with the traditional unlearning problem is that we store the sample-specific knowledge to the deposit module, which is directly vanished in previous unlearning methods.
The deposit module should have two vital characteristics: firstly, it should be withdrawn easily to the recover network with the withdrawal request; Secondly, it should be light-weight to be stored.

To get a better knowledge container, we initialized the deposit module with the pruned original network: 
\begin{equation}
   \mathcal{T}_r\xleftarrow{\text{initialize}} \mathcal{P}rune\big[ \mathcal{T}_0^{(-n)}\big],
\end{equation}
where we use the simple ranking method by calculating the sum of its absolute kernel weights for pruning~\cite{2016Pruning}. The detail of pruning is given in the supplementary.

Here, the deposit module is designed as the pruned version mainly for the following purposes: one is for model efficiency that the light-weight deposit module is more space-saving for storage ($20\%$ parameters of the original network); 
the other is for knowledge filtering that pruning would be better described as `selective knowledge damage'~\cite{Hooker2020WhatDC}, where only the activated filters are kept such that we only deposit the sample-specific knowledge of $\mathcal{D}_r$ rather than the whole knowledge.
Also, similar as $\mathcal{L}_{kp}$, the partial knowledge transfer loss $\mathcal{L}_{pt}$ with the filter $\overline{g}$ is applied to augment this sample-specific knowledge by:
\begin{equation}
    \mathcal{L}_{pt} = \mathcal{L}_{kd}\big(\overline{g}(\frac{z_{\mathcal{T}_r\circ\mathcal{T}^{(n-)}}(x)}{T}),\overline{g}(\frac{z_{\mathcal{T}_{0}}(x)}{T})\big), 
    \label{eq:pt}
\end{equation}
where $\mathcal{L}_{kd}$ and $T$ are previously defined in Eq.~(\ref{eq:kp}) and the logits produced by the deposit module are processed by $\mathcal{T}_r$ and $\mathcal{T}^{(n-)}$. The filter $\overline{g}$ selects the logits that correspond to the class of the deposit set, which transfers the $\mathcal{D}_r$-related knowledge from the original network to the deposit module. 


By minimizing the loss $\mathcal{L}_{pt}$, the sample-specific knowledge is transferred to the deposit module, at the meanwhile we also finetune $\mathcal{T}_r$ to the easy-to-withdraw module, which means that the knowledge is recoverable for the recover network $\widetilde{{\mathcal{T}}}$.
Hence the recovered performance on $\mathcal{D}_r$ is considered ahead in the deposit process, which is to minimize the classification loss of the recover net $\mathcal{L}_{re}$:
\begin{equation}
    \mathcal{L}_{re} = \mathcal{L}_{ce}\big(\widetilde{{\mathcal{T}}}(x),y\big),
    \label{eq:recover}
\end{equation}
where $y$ is the groundtruth label of input $x$. And the deposit module obtained here only works for storing the knowledge, which can't 
be used as a independent prediction model. Thus, $\mathcal{D}_r$ is much more safer form for storage than the original images.  

\subsection{Withdraw Knowledge to Recover Net}
\label{sec:recover}
Once the knowledge has been successfully deposited, the proposed LIRF framework is completed, where the knowledge can be withdrawn directly without any fine-tuning, let alone no need for any data. 

The recover net is re-organized without fine-tuning, which is in the form as:
\begin{equation}
    \widetilde{{\mathcal{T}}}(x)=g\big (\mathcal{T}(x)\big)+ \overline{g}\big ( \mathcal{T}_r\circ\mathcal{T}^{(n-)}(x)\big),
\end{equation}
where the filter functions $g$ and $\overline{g}$ are doing the selection operation, which are the same in Eq.~(\ref{eq:kp}) and Eq.~(\ref{eq:pt}), respectively.
Thus, the overall loss function to update the LIRF framework is:
\begin{equation}
    \mathcal{L}_{all}= \mathcal{L}_{kr}+\lambda_{kp}\mathcal{L}_{kp}+\lambda_{re}\mathcal{L}_{re}+\lambda_{pt}\mathcal{L}_{pt},
\end{equation}
where $\lambda_{kp}$, $\lambda_{pt}$ and $\lambda_{re}$ are the balancing weights. LIRF is trained by minimizing the overall $\mathcal{L}_{all}$ on the deposit set $\mathcal{D}_r$, where the preservation set $\overline{\mathcal{D}}_r$ doesn't participate in the whole process. 

\subsubsection{More discussions.}
Note that the optimization objective $\mathcal{L}_{all}$ of knowledge deposit is different from machine unlearning, which aims at obtaining a target network that approximates the one trained from scratch with data $\overline{ \mathcal{D}}_r$. 
In the proposed LIRF, the knowledge capacity of target net is larger than the network only trained with $\overline{ \mathcal{D}}_r$, for it contains the general knowledge filtered from the delete set $\mathcal{D}_r$. And only the sample-specific knowledge that is privacy-related should be stored in the deposit module.
In the process of withdrawal, the recover network $\widetilde{\mathcal{T}}$ built in Eq.~(\ref{eq:recover}) isn't forced to approach the original network: $\widetilde{\mathcal{T}}\neq\mathcal{T}_0$. Actually the recover network works better than the original network with the existence of full and partial knowledge transfer.

\section{Experiments}

\subsection{Experimental settings}
\textbf{Datasets.}
We use three widely used benchmark datasets for life-long learning, which are CIFAR-10, CIFAR-100 and CUB200-2011 datasets~\cite{Wah2011TheCB}.
For CIFAR-10 and CIFAR-100 datasets, we are using input size of $32\times 32$.
For CUB200-2011 dataset, we are using input size of $256\times 256$.
In the normal knowledge deposit and withdrawal setting, 
the first 30\% of classes are selected for the deposit set, while the rest classes belong to the preservation set.

\noindent
\textbf{Training details.}
We used PyTorch framework for the implementation.
We apply the experiments on the ResNet-18 backbone.
For optimizing the target network and the deposit module, we use stochastic gradient descent with  momentum of 0.9 and learning rate of 0.01 for 20 epochs. 
We employed a standard data augmentation strategy: random crop, horizontal flip, and rotation.
For applying distillation, we set $T=10$ for CIFAR10 dataset and $T=20$ for CUB200-2011 dataset.
For the weights balancing the loss items in $\mathcal{L}_{all}$,
we set $\lambda_{kp}=\lambda_{pt}= \lambda_{re}=10$.
For the normal LIRF setting, the pruning rate is set as $50\%$ and the original network $\mathcal{T}_0$ is divided into 4 blocks, where the last 2 blocks as well as the fc layers are formed as $\mathcal{T}_0^{(n-)}$.

\noindent
\textbf{Evaluation metrics.}
 For evaluation, we need to evaluate the performance of both target net and recover net.
 For recover net, we use the average accuracy for the preservation set (Pre Acc.), the average accuracy for the deposit set (Dep Acc.), the average accuracy for the full set (Avg Acc.).
And for target net, we use the average accuracy for the preservation set (Pre Acc.) and the the average accuracy for the deposit set (Dep Acc.) for the deposit set $\overline{\mathcal{D}}_{r}$.
In addition, following the setting of LWSF~\cite{Shibata2021LearningWS}, we use the harmonic mean (H Mean) of the two standard evaluation measures for life-long learning, which is computed by:
$ H Mean=\frac{2\cdot Pre Acc\cdot F}{Pre Acc+F}$,
where the forgetting measure `F' is computed for the
deposit set by the accuracy drop (decrease) before and after knowledge deposit. For testing the withdrawal performance, all the metrics show better performance with higher values, which are similar for testing the deposit performance on target net, except that `Dep Acc.' is better with lower values.

\subsection{Experimental Results}
\subsubsection{Overall performance.}
Table~\ref{tab:mainacc} shows overall performance of knowledge deposit (target network) and withdrawal (recover network) on CIFAR-10 and CUB200-2011 datasets.
Besides, we compare the proposed LIRF with the `Independent' networks independently trained on the two sub datasets (preservation set and deposit set) and the `Original' network $\mathcal{T}_0$ trained on the full dataset. From Table~\ref{tab:mainacc}, several observations are obtained:
\begin{itemize}
    \item In the original network, the accuracy on preservation set (`Pre Acc.') is higher than the performance trained dependently (`93.77' vs `92.92' on CIFAR-10), which means that there exits the positive knowledge transfer from the deposit $\mathcal{D}_r$ to the preservation $\overline{\mathcal{D}}_r$. Thus, it is necessary to partial transfer the general knowledge to the preserved tasks.
    \item As the accuracy for randomly predicting on CIFAR-10 and CUB200-2011 is $10\%$ and $0.5\%$, respectively,  the `Pre Acc.' while depositing decreases to $15\%$ and $1.18\%$. This large accuracy drop demonstrates the logit-level forgetting of the deposit set in the target net $\mathcal{T}$.
    \item The recover network gains higher accuracy on both the preservation set and the deposit set than on the original network, which proves the knowledge has been augmented in LIRF with the partial and full knowledge transfer, which demonstrates the discussions in Sec.~\ref{sec:recover},
\end{itemize}

\begin{table}[t]
\caption{Experimental results of the proposed LIRF on CIFAR-10 dataset and CUB200-2011 dataset. For each dataset, we randomly select $30\%$ of classes for deposit (Dep Set), while the rest is kept in the preservation set (Pre Set).}
\centering
\label{tab:mainacc}
\begin{tabular}{p{21mm}|p{16mm}<{\centering}|p{18mm}<{\centering}|p{18mm}<{\centering}|p{18mm}<{\centering}|p{18mm}<{\centering}}
\toprule
 Dataset& Metrics  &Independent& Original& Deposit & Withdrawal \\\hline\hline
CIFAR-10& Pre Acc.$\uparrow$   &  92.92   &  93.77  &   93.41 & 94.56          \\
CIFAR-10& Dep Acc.  &  96.61  & 94.60  &  15.00 & 97.92        \\
CIFAR-10& F$\uparrow$   &  - & 0  &  79.60 &-          \\
CIFAR-10& H Mean $\uparrow$ &  -  & 0  &   85.95&-         \\
CIFAR-10& Avg Acc. $\uparrow$  &  94.02  &  94.06 &   - & 95.57    \\\hline\hline
CUB200-2011& Pre Acc.$\uparrow$   &   48.15  & 50.33 & 51.64& 53.21       \\
CUB200-2011& Dep Acc.  &   52.73 &  48.60 & 1.18  & 55.89   \\
CUB200-2011& F$\uparrow$  &  -  &  0  &   47.42&- \\
CUB200-2011& H Mean $\uparrow$ &  -  &0   & 49.44 &-      \\
CUB200-2011& Avg Acc. $\uparrow$ & 49.52  & 49.81& - & 54.01 \\
\bottomrule
\end{tabular}
\end{table}

\begin{table}[t]
\centering
\caption{Experimental results of the ablation study on the proposed LIRF framework.}
\label{tab:ablation}
\begin{tabular}{c|cc|ccc}
\toprule
\multicolumn{1}{c|}{\multirow{2}{*}{Method}} & \multicolumn{2}{c|}{\#Target Net} & \multicolumn{3}{c}{\#Recover Net} \\
\multicolumn{1}{c|}{}&  Pre Acc. $\uparrow$   & Dep Acc. $\downarrow$  & Pre Acc. $\uparrow$      & Dep Acc.  $\uparrow$ & Avg Acc.  $\uparrow$    \\ \hline\hline
Scratch Train & 92.92 &96.61 & 93.77&  94.60&94.06\\
IL Train &  92.92 &96.61  &90.87 &\textbf{98.37}&93.12   \\
$\mathcal{L}_{kr}$,$\mathcal{L}_{kp[w/o\mathcal{L}_{at}]}$ & 93.38 & 15.55& -&-&-\\
$\mathcal{L}_{kr},\mathcal{L}_{kp[w/o\mathcal{L}_{at}]},\mathcal{L}_{pt},\mathcal{L}_{re}$ &  93.25&14.81&94.26&97.03&95.09\\ 
$\mathcal{L}_{kr},\mathcal{L}_{kp},\mathcal{L}_{pt},\mathcal{L}_{re}$ & 93.25 &15.29 &94.33 &97.07&95.15\\ 
$\mathcal{L}_{kr},\mathcal{L}_{kp},\mathcal{L}_{pt},\mathcal{L}_{re}$+Prune& \textbf{93.42}&\textbf{14.15} &\textbf{94.55}&97.67&\textbf{95.49}\\\bottomrule
\end{tabular}
\end{table}

\begin{figure}[t]
\centering
\includegraphics[scale = 0.48]{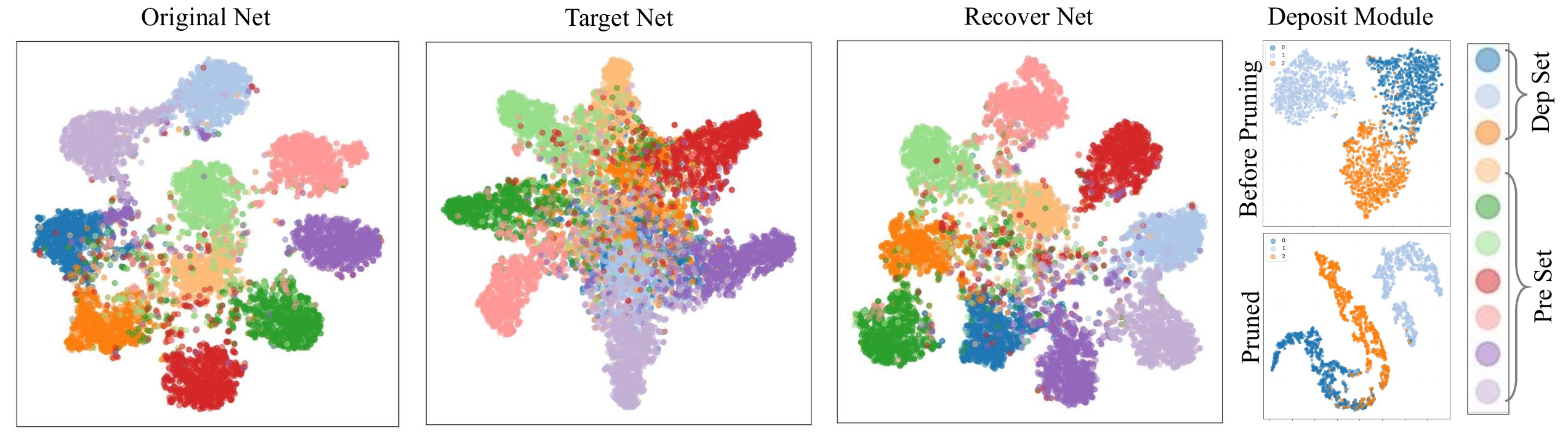}
\caption{t-SNE plots of the features obtained from the last layer of the network are shown. Each color in the t-SNE plot represents one category, where 3 categories are deposited and the rest 7 categories are in the preservation set (best viewed in color).
}
\label{fig:tsne}
\end{figure}

\subsubsection{Sensitive analysis of LIRF.}
Here we give a deeper analysis of the proposed LIRF by the ablation study of each loss items in $\mathcal{L}_{all}$.
The comparative results are applied on, 
`Scratch Train': train each net with the corresponding set from scratch; 
`IL Train': train the target net from scratch and train recover net with KD loss in the incremental learning setting; 
$\mathcal{L}_{kr},\mathcal{L}_{kp},\mathcal{L}_{pt},\mathcal{L}_{re}$ are the loss items defined in the LIRF framework;
$\mathcal{L}_{kp[w/o\mathcal{L}_{at}]}$ is the loss without the attention distillation $\mathcal{L}_{at}$;
`Prune' denotes the pruning operation to initialize the deposit module.
The experimental results are displayed in Table~\ref{tab:ablation}. 
As can be observed from the table:
(1) The full setting with all the loss functions and the pruning strategy achieves almost the best on each metrics. 
(2) The attention loss item $\mathcal{L}_{at}$ would not affect the accuracy a lot (`row 4' and `row 5'), but it is of vital importance to prevent the information leakage, which is discussed in the following experiment.
(3) The pruning strategy on the deposit module is proved to be effective since the pruned deposit module can be withdrawn to recover net with the best Avg Acc.(`95.49').

The visualization of the t-SNE plots is depicted in Fig.~\ref{fig:tsne}, where the features
on the final layer of original net, target net and recover net are visualized. As is shown in the figure, both the original net and the recover net can produce discriminative features on all the 10 categories. And for the target net where the sample-specific knowledge of the deposit set is removed, the visualization proves that the target net produces highly discriminative features for the preservation set while vanishing the predicting capacity for the deposit set.
And for visualization the t-SNE plots of the deposit module, we construct a network as $\mathcal{T}_r\circ\mathcal{T}_0^{(n-)}$. As can been seen in the right part of the figure, the pruned deposit module produces more `narrow' features, which are thought as the sample-specific knowledge we want to deposit, proving the `selective knowledge damage' scheme we mentioned in Sec.~\ref{sec:deposit}.

\begin{table}[t]
\centering
\caption{Experimental results of the knowledge transferability to downstream networks. This experiment is conducted on the CIFAR-10 dataset.}
\label{tab:distillation}
 \resizebox{1.0\textwidth}{!}{
\begin{tabular}{c|c|cc|cc|cc}
\toprule
\multicolumn{1}{c|}{\multirow{2}{*}{Student}}&\multicolumn{1}{c|}{\multirow{2}{*}{Distillation}}& \multicolumn{2}{c|}{\#Original}  & \multicolumn{2}{c|}{\#Target (w/0 $\mathcal{L}_{at}$)} &\multicolumn{2}{c}{\#Target (w $\mathcal{L}_{at}$)}\\
\multicolumn{1}{c|}{}& & Pre Acc.  & Dep Acc. & Pre Acc.$\uparrow$& Dep Acc.$\downarrow$& Pre Acc.$\uparrow$& Dep Acc.$\downarrow$\\ \hline\hline
CNN&Logit-based&85.38&86.15&85.97\up{(+0.59)}&84.26\down{(-1.89)}&85.70\up{(+0.32)}&82.75\down{(-3.40)}\\
CNN&AT-based&85.27&85.94&86.01\up{(+0.74)}&85.72\down{(-0.22)}&85.54\up{(+0.27)}&81.83\down{(-4.11)}\\
ResNet18&Logit-based&94.26&95.73&94.55\up{(+0.29)}&92.70\down{(-3.03)}&94.64\up{(+0.38)}&91.49\down{(-4.24)}\\
ResNet18&AT-based& 94.09&95.24&94.15\up{(+0.06)}&94.61\down{(-0.63)}& 93.85\down{(-0.24)}&88.76\down{(-6.48)}\\\bottomrule
\end{tabular}
}
\end{table}

\subsubsection{Knowledge transferability to downstream networks.}
We use two evaluations to prove the success of sample-specific knowledge removal in the target net:
One is the accuracy drop on the deposit set, which has been proved in the former experiments;
The other is the knowledge transferability of the deposit set from the target network to downstream networks, which test the knowledge leakage risk by knowledge distillation.
We have conducted the logit-based distillation (KL-divergence loss of the output logits) and the attention-based distillation (MSE loss of the attention maps of the intermediate features). The results are displayed in Table~\ref{tab:distillation}, where we choose the plain CNN and the ResNet-18 as the student. And we have also evaluated the necessity of the loss item $\mathcal{L}_{at}$ in $\mathcal{L}_{kr}$. Note that the groudtruth label of the training data is included for training with distillation, the accuracy wouldn't drop largely even when the knowledge is nontransferable.
Thus, from the table, we observe that:
(1) The knowledge transferability on the preservation set is guaranteed on both the original and the target networks, which is slightly influenced by target net with $\mathcal{L}_{at}$ distilled in the attention-based way;
(2) When training the LIRF framework with loss item $\mathcal{L}_{at}$, the knowledge for the deposit set is hard to be distilled both in the attention-based and the logit-based way. It is much safer with $\mathcal{L}_{at}$, since without it, the knowledge of $\mathcal{D}_r$ is likely to be leaked through distillation in the attention-based manner. 
The privacy protection on the deposit set is further evaluated in the supplementary tested by the data-free distillation. 

\subsubsection{The influence of the scale of the deposit module.}
\begin{figure}[t]
\centering
\includegraphics[scale = 0.36]{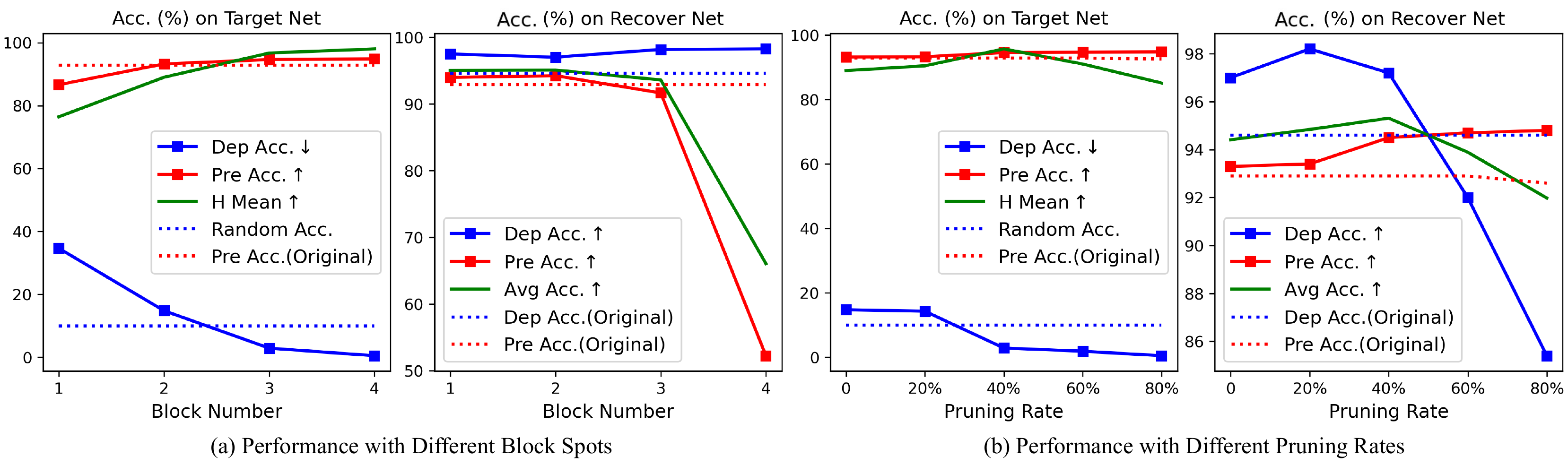}
\caption{The performance on knowledge deposit and withdrawal with different division block numbers and pruning rates.
}
\label{fig:size}
\end{figure}

There are two factors corresponding to the scale of the deposit module $\mathcal{T}_r$: one is the block number used to divide the original network;
the other is the pruning rate.
The influence on this two factors is depicted in Fig.~\ref{fig:size}.
When the division block number $n$ increases, the size of the deposit module becomes larger and the fully transferred part of the original network ($\mathcal{T}_0^{(n-)}$) becomes smaller. In this way, the deposit accuracy on the target network (`Dep Acc.' in the first sub figure) and the average accuracy on the recover network (`Avg Acc.' in the second sub figure) drop due to the less knowledge directly transferred from $\mathcal{T}_0^{(n-)}$ to $\mathcal{T}$, which is also completely transferred back while recovering.
Considering the performance on both target net and recover net, dividing at $n=2$ and $n=3$ satisfies the demand, and we choose $n=2$ for smaller knowledge deposit storage.
When the pruning rate increases (the percent of filters to be pruned out), the size of the deposit module becomes smaller, which doesn't influence the deposit performance largely (the third sub figure). The average accuracy on the recover network(`Avg Acc.' in the forth sub figure) increases at first due to selective knowledge damage on the deposit module, but drops at last due to the limit size for knowledge storage. So the pruning rate near $50\%$ is a better choice.

\begin{table}[t]
\caption{Comparative results of incremental learning and unlearning on CIFAR-100. Each column represents a different number of classes per incremental step.}
\label{tab:incremental}
\centering
\begin{tabular}{l|cc|cc|cc}
\toprule
 & \multicolumn{2}{c|}{\# Task:2, \# Class:50} & \multicolumn{2}{c|}{\# Task:5, \# Class:20} & \multicolumn{2}{c}{\# Task:20, \# Class:5} \\
 & H $\uparrow$ & (A$\uparrow$, F$\uparrow$)   &H $\uparrow$ & (A$\uparrow$, F$\uparrow$)  & H $\uparrow$ & (A$\uparrow$, F$\uparrow$)  \\\hline\hline
Baseline & 55.87 & (55.21,56.55) & 51.79 &(39.66,74.62) &37.88 &(25.41,74.41)  \\
LwF&9.02 &(74.69,4.80) &17.23 &(79.05,9.67) &22.50& (80.74,13.07)\\
LwF*& 54.64 &(76.44,42.52) &68.24 &(81.32,58.79)& 63.62 &(82.29,51.85)\\
EWC& 58.58& (56.73,60.55) &48.57 &(36.54,72.42)& 34.91& (23.07,71.70)\\
EWC*& 57.17& (56.25,58.13)& 49.61& (36.58,77.08)& 36.90& (23.68,83.52)\\
EWC*+LwF*& 53.51& (77.11,40.98)& 67.64 &(81.20,57.96)& 69.17& (74.11, 64.85)\\
MAS& 55.44& (54.42,56.49)& 47.46 &(34.89,74.17)& 35.26& (23.25,72.96)\\
MAS+LwF*& 56.54& (76.85,44.72)& 66.35 &(81.83,55.79)& 70.83& (74.63,67.41)\\
LWSF&70.08 &(74.89,65.84)& \textbf{73.21}& (72.61,73.83)& 71.63& (68.56,75.00)\\\hline
LIRF &\textbf{77.69}&(79.24,76.19)&{73.08} &({78.24}, {68.56})& \textbf{79.48}&(80.41,78.57)\\
 \bottomrule
\end{tabular}
\end{table}

\subsubsection{Comparing with incremental learning and unlearning.}
The proposed LIRF can be also conducted on incremental learning task and the machine unlearning task. 
Here we tested LIRF on these two tasks, following the setting of LWSF~\cite{Shibata2021LearningWS}, whose task is to unlearn several classes of samples while dealing with the new classes. The individual experiments on incremental learning and machine unlearning are given in the supplementary.
To begin with, we train the original network with the full dataset and then deposit each sub class into a deposit module set, then withdraw each in each incremental step.
Table.~\ref{tab:incremental} shows the comparative results of all
the methods, which include: `Baseline' (trained only with the classification loss),
`LwF'~\cite{Li2016LearningWF}, `EWC'~\cite{kirkpatrick2017overcoming}, `MAS'~\cite{Aljundi_2018_ECCV}, LWSF~\cite{Shibata2021LearningWS} and the modified `LwF*' and `EWC*' that are modified to enable partial forgetting by the work~\cite{Shibata2021LearningWS}. The metrics of `H Mean' (H), `Pre Acc' (A) and `F' are averaged until the last incremental step. The proposed LIRF works almost the best among all the listed methods, especially on the incremental performance (`A') which is due to partial and fully knowledge transfer in the framework.



\section{Conclusions}
In this paper, we study a novel life-long learning task,
recoverable knowledge forgetting.
Unlike prior life-long learning tasks that either aim to
prevent the forgetting of old knowledge or
delete specified knowledge for good,
the investigated  setting enables
flexible knowledge extraction and inserting,
which in turn largely enriches the 
user control and meanwhile ensures network IP protection.
To this end, we introduce a dedicated approach,
termed as LIRF, where the innovative operation
of knowledge deposit and knowledge withdrawal 
are proposed. During deposit, 
the sample-specific knowledge that may lead to privacy leakage
is extracted from original network and 
maintained in the deposit module.
Whenever needed, the deposited knowledge can be readily 
withdrawn to recover the original model.
Experimental results demonstrate the effectiveness of the 
proposed LIRF under various settings, including 
incremental learning and machine unlearning.

\section*{Acknowledgements}
This work is supported by 
NUS Advanced Research and Technology Innovation Centre~(Project Reference: ECT-RP2), Centre for Advanced Robotics Technology Innovation~(CARTIN) of 
Singapore, NUS Faculty Research Committee~(WBS: A-0009440-00-00),
National Natural Science Foundation of China (No.62002318), Zhejiang Provincial Science and Technology Project for Public Welfare (LGF21F020020) and Ningbo Natural Science Foundation (202003N4318).
Xinchao Wang is the corresponding author.

\clearpage
%
%

\end{document}